\documentclass[11pt,a4paper]{article}
\usepackage[hyperref]{acl2019}
\usepackage{times}
\usepackage{latexsym}

\usepackage{url}

\usepackage{amsmath,amssymb,amsthm}
\usepackage{color}
\usepackage{subcaption,siunitx,booktabs}
\usepackage{makecell}
\usepackage{arydshln}
\usepackage{soul}
\usepackage{xcolor}
\usepackage{xspace}

\usepackage{graphicx}

\DeclareMathOperator{\softmax}{softmax}

\newcolumntype{?}{!{\vrule width 1.5pt}}

\aclfinalcopy 

\newcommand{\newvec}[1]{\mathbf{#1}}
\newcommand{\esqe}{{\sc ESQE}\xspace}

\title{This Email Could Save Your Life: Introducing the Task of Email Subject Line Generation}

\author{Rui Zhang \thanks{\quad Work done during the internship at Grammarly.} \\ Yale University \\ {\tt r.zhang@yale.edu} \And Joel Tetreault \\ Grammarly \\ {\tt joel.tetreault@grammarly.com}}

\date{}
\begin{document}
\maketitle
\begin{abstract}
Given the overwhelming number of emails, an effective subject line becomes essential to better inform the recipient of the email's content.
In this paper, we propose and study the task of \textit{email subject line generation}: automatically generating an email subject line from the email body.
We create the first dataset for this task and find that email subject line generation favor extremely abstractive summary which differentiates it from news headline generation or news single document summarization.
We then develop a novel deep learning method and compare it to several baselines as well as recent state-of-the-art text summarization systems.
We also investigate the efficacy of several automatic metrics based on correlations with human judgments and propose a new automatic evaluation metric.
Our system outperforms competitive baselines given both automatic and human evaluations.
To our knowledge, this is the first work to tackle the problem of effective email subject line generation.
\end{abstract}

\vspace{-4mm}
\section{Introduction}
\vspace{-1mm}
Email is a ubiquitous form of online communication.
An email message consists of two basic elements: an \textit{email subject line} and an \textit{email body}.
The subject line, which is displayed to the recipient in the list of inbox messages, should tell what the email body is about and what the sender wants to convey.
An effective email subject line becomes essential since it can help people manage a large number of emails.
Table \ref{tab:example} shows an email body with three possible subject lines.

\begin{table}[t!]
\centering
\resizebox{\columnwidth}{!}{
\begin{tabular}{?p{9cm}?}
\Xhline{4\arrayrulewidth}
\textbf{Email Body:} Hi All, I would be grateful if you could get to me today via email a job description for your current role.
I would like to get this to the immigration attorneys so that they can finalise the paperwork in preparation for INS filing once the UBS deal is signed.
Kind regards, \\
\textbf{Subject 1:} Current Job Description Needed \textit{(COMMENT: This is good because it is both informative and succinct.)} \\
\textbf{Subject 2:} Job Description \textit{(COMMENT: This is okay but not informative enough.)} \\
\textbf{Subject 3:} Request \textit{(COMMENT: This is bad because it does not contain any specific information about the request.)} \\
\Xhline{4\arrayrulewidth}
\end{tabular}
}
\caption{An email with three possible subject lines.
}
\vspace{-4mm}
\label{tab:example}
\end{table}

There have been several research tracks around email usage.
 While much effort has been focused on email summarization \cite{muresan2001combining,nenkova2003facilitating,rambow2004summarizing}, email keyword extraction and action detection \cite{turney2000learning,lahiri2017keyword,lin2018actionable}, and email classification \cite{prabhakaran2014predicting,alkhereyf2017work}, to our knowledge there is no previous work on generating email subjects.
In this paper, we propose the task of Subject Line Generation (SLG): automatically producing email subjects given the email body.
While this is similar to email summarization, the two tasks serve different purposes in the process of email composition and consumption.
A subject line is required when the sender writes the email, while a summary is more useful for long emails to benefit the recipient.
An automatically generated email subject can also be used for downstream applications such as email triaging to help people manage emails more efficiently.
Furthermore, while being similar to news headline generation or news single document summarization, email subjects are generally much shorter, which means a system must have the ability to summarize with a high compression ratio (Table \ref{tab:data}).
Therefore, we believe this task can also benefit other highly abstractive summarization such as generating section titles for long documents to improve reading comprehension speed and accuracy.

\begin{table*}[ht!]
\centering
\resizebox{\textwidth}{!}{
\begin{tabular}{lcccc}
\Xhline{4\arrayrulewidth}
Dataset                      & domain    & docs (train/val/test)     & avg doc words & avg summary words \\ \hline
CNN \cite{cheng2016neural}   & News      & 90,266/1,220/1,093        & 760           & 46                \\
XSum \cite{narayan2018dont}  & News      & 204,045/11,332/11,334     & 431           & 23                \\
Gigaword News Headline \cite{rush2015neural}      & News      & 3,799,588/394,622/381,197 & 31            & 8  \\
Annotated Enron Subject Line Corpus & Business/Personal  & 14,436/1,960/1,906        & 75            & 4                 \\
\Xhline{4\arrayrulewidth}
\end{tabular}
}
\caption{Annotated Enron Subject Line Corpus compared with other datasets.}
\vspace{-3mm}
\label{tab:data}
\end{table*}

To introduce the task, we build the first dataset, Annotated Enron Subject Line Corpus (AESLC), by leveraging the Enron Corpus \cite{klimt2004enron} and crowdsourcing.
Furthermore, in order to properly evaluate the subject, we use a combination of automatic metrics from the text summarization and machine translation fields, in addition to building our own regression-based Email Subject Quality Estimator (\esqe).
Third, to generate effective email subjects, we propose a method which combines extractive and abstractive summarization using a two-stage process by \textit{Multi-Sentence Selection and Rewriting with Email Subject Quality Estimation Reward}.
The multi-sentence extractor first selects multiple sentences from the input email body.
Extracted sentences capture salient information for writing a subject such as named entities and dates.
Thereafter, the multi-sentence abstractor rewrites multiple selected sentences into a succinct subject line while preserving key information.
For training the network, we use a multi-stage training strategy incorporating both supervised cross-entropy training and reinforcement learning (RL) by optimizing the reward provided by the \esqe model.

Our contributions are threefold: (1) We introduce the task of email subject line generation (SLG) and build a benchmark dataset AESLC.\footnote{dataset available at \url{https://github.com/ryanzhumich/AESLC}}  (2) We investigate possible automatic metrics for SLG and study their correlations with human judgments.  We also introduce a new email subject quality estimation metric (\esqe). (3) We propose a novel model to generate email subjects. Our automatic and human evaluations demonstrate that our model outperforms competitive baselines and approaches human-level quality.

%
%

\section{Annotated Enron Subject Line Corpus}
To prepare our email subject line dataset, we use the Enron dataset \cite{klimt2004enron} which is a collection of email messages of employees in the Enron Corporation.
We use Enron because it can be released to the public and it contains business and personal type emails for which the subject line is already well-defined and useful.
As shown in Table \ref{tab:data}, email subjects are typically much shorter than summaries generated in previous news datasets.
While being similar to news headline generation \cite{rush2015neural}, email subject generation is also more challenging in the sense that it deals with different types of email subjects while the first sentence of a news article is often already a good headline and summary.

\subsection{Data Preprocessing}
The original Enron dataset contains 517,401 email messages from 150 user mailboxes.
To extract body and subject pairs from the dataset, we take all messages from the inbox and sent folders of all mailboxes.
We then perform email body cleaning, email filtering, and email de-duplication.

We first remove any content from the email body that has not been written by the author of the email.
This includes automatically appended boilerplate material such as advertisements, attachments, legal disclaimers etc.
Since we are interested in emails with enough information to generate meaningful subjects, we only keep emails with at least 3 sentences and 25 words in the email body.
Furthermore, to ensure that the email subject truly corresponds to the content in the email body, we only take the first email of a thread and exclude replies or forward emails.
So we filter out follow up messages which contain ``Original Message" section in the email body or have subject lines starting with ``RE:" (reply-to messages) or ``FW:" (forward messages).
Finally, we observe that the same message can be sent to multiple recipients so we remove duplicate emails to make sure there is no overlap between the train and test set.
We only keep the subject and body while other information such as the sender/recipient identity can be incorporated in future work.

\subsection{Subject Annotation}
We noted that using only the original subject lines as references may be problematic for automatic evaluation purposes.
First, there can be many different valid, effective subject lines for the same email, yet the original email subject is only one of them. This is similar to why automatic machine translation evaluation often relies on multiple references.
Second, the email subject may be too general or too vague when the sender does not put that much effort into writing.
Third, the sender may assume some shared knowledge with the recipient so that the email subject contains information that cannot be found in the email body.

To address the issues above, we ask workers on Amazon Mechanical Turk to read Enron emails in our dev and test sets and write an appropriate subject line.
Each email is annotated with 3 subject lines from 3 different annotators.
For quality control, we manually review and reject improper email subjects such as empty subject lines, subject lines with typos, and subject lines that are too general or too vague, e.g., ``Update", ``Schedule", ``Attention to Detail" because they contain no body-specific information and can be applied generically to many emails.  
We found that while three annotations are different, they often contain common keywords.
To further quantify the variation among human annotations, we compute ROUGE-L F1 scores for each pair of annotations: 34.04, 33.38, 34.26.

\section{Our Model}
\begin{figure*}[t]
  \centering
  \includegraphics[width=.8\textwidth]{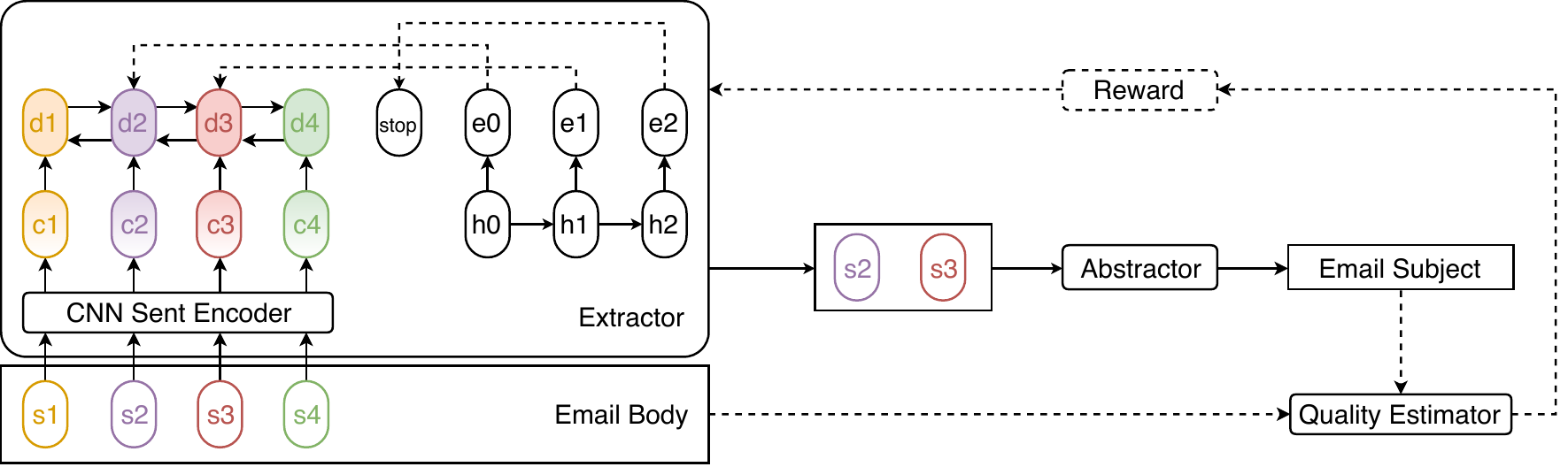}
  \caption{Our model architecture. In this example, the input email body consists of four sentences from which the extractor selects the second and the third. The abstractor generates an email subject from the selected sentences. The quality estimator provides rewards by scoring the subject against the email body.}
  \vspace{-4mm}
  \label{fig:model}
\end{figure*}

Our model is illustrated in Figure \ref{fig:model}.
Based on recent progress in news summarization \cite{chen2018fast}, our model generates email subjects in two stages:
(1) The extractor selects multiple sentences containing salient information for writing a subject (\S\ref{sec:extractor}).
(2) The abstractor rewrites multiple selected sentences into a succinct subject line while preserving key information (\S\ref{sec:abstractor}).

We employ a multi-stage training strategy (\S\ref{sec:training}) including a Reinforcement Learning (RL) phase because of its usefulness for text generation tasks \cite{ranzato2016sequence,bahdanau2017actor} to optimize the non-differentiable metrics such as ROUGE and METEOR.
However, unlike ROUGE for summarization or METEOR for machine translation, there is no available automatic metric designed for email subject generation.
Motivated by recent work on regression-based metrics for machine translation \cite{shimanaka2018metric} and dialog response generation \cite{lowe2017towards}, we build a neural network (ESQE) to estimate the quality of an email subject given the email body (\S\ref{sec:esqe}).
The estimator is pretrained and fixed during RL training phase to provide rewards for the extractor agent.

While our model is based on \newcite{chen2018fast}, they assume that there is a one-to-one relationship between the summary sentence and the document sentence: every summary sentence can be rewritten from exactly one sentence in the document.
They also use ROUGE to make extraction labels and to provide rewards in their RL training phase.
In contrast, our model extracts multiple sentences and rewrites them together into a single subject line.
We also use word overlap to make extraction labels and use our novel \esqe as a reward function.

\subsection{Multi-sentence Extractor}
\label{sec:extractor}
For the first stage, we need to select multiple sentences from the email body which contain the necessary information for writing a subject.
This task can be formulated as a sequence-to-sequence learning problem where the output sequence corresponds to the position of ``positive" sentences in the input email body.
Therefore, we use a pointer network \cite{vinyals2015pointer} to first build hierarchical sentence representations during encoding and then extract ``positive" sentences during decoding.

Suppose our input is an email body $D$ which consists of $|D|$ sentences:
\begin{equation*}
  D = [d_1,d_2,\dots,d_j,\dots,d_{|D|}] 
\end{equation*}
We first use a temporal CNN \cite{kim2014conv} to build individual sentence representations.
For each sentence, we feed the sequence of its word vectors into 1-D convolutional filters with various window sizes.
We then apply ReLU activation and then max-over-time pooling.
The sentence representation is a concatenation of activations from all filters
\begin{equation}
  \newvec{c}_j = \textsc{CNN}(d_j), j=1,\dots,|D|
\end{equation}

Then we use a bidirectional LSTM \cite{hochreiter1997long} to capture document-level inter-sentence information over CNN outputs:
\begin{align}
\label{eq:bilstm}
\begin{split}
\overrightarrow{\newvec{d}}_j &= \textsc{LSTM}^{\text{forward}}(\overrightarrow{\newvec{d}}_{j-1},\newvec{c}_j) \\
\overleftarrow{\newvec{d}}_j &= \textsc{LSTM}^{\text{backward}}(\overleftarrow{\newvec{d}}_{j+1},\newvec{c}_j) \\ 
\newvec{d}_j &= [\overrightarrow{\newvec{d}}_j,\overleftarrow{\newvec{d}}_j] \\
\end{split}
\end{align}

For sentence extraction, another LSTM as decoder outputs one ``positive" sentence at each time step $t$.
Denoting the decoder hidden state as $\newvec{h}^{t}$, we choose a ``positive" sentence from a 2-hop attention process.
First, we build a context vector $\newvec{e}^{t}$ by attending all $\newvec{d}_j$:
\begin{align}
\label{eq:context_vec}
\begin{split}
\hat{\alpha}_{j}^{t}    & = \newvec{v}_{e}^{\intercal}\tanh(\newvec{W}_{e}\newvec{d}_j + \newvec{U}_{e}\newvec{h}^{t}) \\
\alpha^{t}   & = \softmax(\hat{\alpha}^{t}) \\
\newvec{e}^{t} & = \sum_{j}\alpha^{t}_{j}\newvec{W}_{e}\newvec{d}_j\\
\end{split}
\end{align}
Then, we get an extraction probability distribution $o^{t}$ over input sentences:
\begin{align}
\label{eq:action}
\begin{split}
\hat{o}_{j}^{t}  & = \newvec{v}_{o}^{\intercal}\tanh(\newvec{W}_{o}\newvec{d}_j + \newvec{U}_{o}\newvec{e}^{t}) \\
P(o^{t}|o^{1},o^{2},\dots,o^{t-1}) & = \softmax(\hat{o}^{t})\\
\end{split}
\end{align}
where $\{\newvec{v},\newvec{W},\newvec{U}\}$ are trainable parameters.

We also add a trainable ``stop" vector with the same dimension as the sentence representation.
The decoder can choose to stop by pointing to this ``stop" sentence.

\vspace{-1mm}
\subsection{Multi-sentence Abstractor}
\vspace{-1mm}
\label{sec:abstractor}
In the second stage, the abstractor takes the selected sentences from the extractor and rewrites them into an email subject.
We implement the abstractor as a sequence-to-sequence encoder-decoder model with the bilinear multiplicative attention \cite{luong2015effective} and copy mechanism \cite{see2017get}.
The copy mechanism enables the decoder to copy words directly from the input document, which is helpful to generate accurate information verbatim even for out-of-vocabulary words.

\vspace{-1mm}
\subsection{Email Subject Quality Estimator}
\vspace{-1mm}
\label{sec:esqe}
Since there is no established automatic metric for SLG, we build our own Email Subject Quality Estimator (\esqe).
Given an email body $D$ and a potential subject for the subject $s$, our quality estimator outputs a real-valued Subject Quality score $\textsc{SQ}(D,s)$.
The email subject and the email body are fed to a temporal CNN.
\begin{align}
\label{eq:estimator}
\begin{split}
\newvec{D} = \textsc{CNN}(D), \newvec{s} = \textsc{CNN}(s) \\
\end{split}
\end{align}
We concatenate the output of CNNs as the email body and subject pair representation.
Then, a single layer feed-forward neural net follows to predict the quality score from the representation.
\begin{align}
\label{eq:sq}
\begin{split}
\textsc{SQ}(D,s) & = \textsc{FFNN}([\newvec{D},\newvec{s}]) \\
\end{split}
\end{align}

To train the estimator, we collect human evaluations on 3,490 email subjects.  
In order to expose the estimator to both good and bad examples, 2,278 of the 3,490 are the original subjects and the remaining 1,212 subjects are generated by an existing summarization system.
Each subject has 3 human evaluation scores (the same human evaluation as explained in \S\ref{sec:metric}) and we train our estimator to regress the average.

The inter-annotator agreement is 0.64 by Pearson's $r$ correlation.
Even though there is no value range restriction for the estimator output, we found the scores returned by our \esqe after training are bounded from 0.0 to 4.0.

\subsection{Multi-Stage Training}
\label{sec:training}
\textbf{Supervised Pretraining.} 
We pretrain the extractor and the abstractor separately using supervised learning.
To this end, we first create ``proxy" sentence labels by checking word overlap between the subject and the body sentence.
For each sentence in the body, we label it as ``positive" if there is some token overlap of non-stopwords with the subject, negative otherwise.
The multi-sentence extractor is trained to predict ``positive" sentences by minimizing the cross-entropy loss.
For the multi-sentence abstractor, we create training examples by pairing the ``positive" sentences and the original subject in the training set.
Then the abstractor is trained to generate the subject by maximizing the log-likelihood.
\\\noindent
\textbf{RL Training for Extractor.}
To formulate this RL task at this stage, we treat the extractor as an agent, while the abstractor is pretrained and fixed.
The \esqe provides the reward by judging the output subject.
At each time step $t$, it observes a state $s_t = (D,d_{o^{t-1}})$, and samples an action $a_t$ to pick a sentence from the distribution in Equation \ref{eq:action}:
\begin{equation}
    a_t \sim \pi_{\theta}(s_t,a_t=j) = P(o^{t}=j)
\end{equation}
where $\pi_{\theta}$ denotes the policy network described in Section \ref{sec:extractor} with a set of trainable parameters $\theta$.
The episode is finished in $T$ actions until the extractor picks the ``end-of-extraction" signal.
Then, the abstractor generates a subject from the extracted sentences and the quality estimator calculates the score.
The quality estimator is the reward received by the extractor:
\begin{equation}
    r(a_{1:T}) = \textsc{SQ}(D,s)
\end{equation}
For training, we maximize the expected reward:
\begin{equation}
    \mathcal{L}(\theta) = \mathbb{E}_{a_{1:T}\sim\pi_{\theta}}[r(a_{1:T})]
\end{equation}
with the following gradient given by the REINFORCE algorithm \cite{williams1992simple}:
\begin{align}
\label{eq:rl}
\begin{split}
    \nabla_{\theta}\mathcal{L}(\theta) = \mathbb{E}_{\pi_{\theta}}[\nabla_{\theta}\log \pi_{\theta} (r-b)] \\
    \approx \sum_{t=1}^{T} \nabla_{\theta} \log \pi_{\theta}(s_t,a_t) (r(a_{1:T})-b_{t})\\
\end{split}
\end{align}
$b_{t}$ is the baseline reward introduced to reduce the high variance of gradients.
The baseline network has the same architecture as the decoder of the extractor.
But it has another set of trainable parameters $\theta_{b}$ and predicts the reward by minimizing the following mean squared error:
\begin{equation}
    \mathcal{L}(\theta_{b}) = (b_t - r)^2
\end{equation}

\begin{table*}[t!]
\begin{subtable}{1\textwidth}
\resizebox{\textwidth}{!}{
\begin{tabular}{l|cccc|cccc}
\Xhline{4\arrayrulewidth}
                          & \multicolumn{4}{c|}{Dev} & \multicolumn{4}{c}{Test}\\
                          & R-1 & R-2 & R-L & METEOR & R-1 & R-2 & R-L & METEOR \\ \Xhline{4\arrayrulewidth}
LEAD-2                    & 11.28 & 4.61 & 10.48 & 10.76 & 11.00 & 4.33 & 10.20 & 11.27$^\ast$        \\
TextRank                  & 11.12 & 3.75 & 10.15 & 9.19  & 11.32 & 3.88 & 10.14 & 10.64$^\ast$       \\
LexRank                   & 13.02 & 4.96 & 11.89 & \underline{10.84} & 12.46 & 4.62 & 11.37 & \underline{11.56}$^\ast$ \\

\newcite{shang2018unsupervised}    & 10.56 & 3.28 & 9.92 & 6.17 & 10.40 & 3.09 & 9.77 & 6.15       \\ 
\newcite{see2017get}      & 18.02 & \underline{5.73} & 16.63 & 10.83 & \underline{17.02} & \underline{5.45} & 15.78 & 10.31       \\
\newcite{paulus2018deep}  & 14.08 & 5.09 & 13.36 & 9.07 & 13.49 & 4.55 & 12.83 & 8.65       \\
\newcite{hsu2018unified}  & 16.59 & 4.67 & 15.12 & \textbf{13.22}$^\ast$ & 15.75 & 4.54 & 14.41 & \textbf{12.49}$^\ast$       \\
\newcite{narayan2018dont} & 13.52 & 3.27 & 13.33 & 4.64 & 12.60 & 3.09 & 12.52 & 4.66 \\ 
Our System                & \textbf{25.41} & \textbf{11.34} & \textbf{25.07} & 9.83 & \textbf{23.67} & \textbf{10.29} & \textbf{23.44} & 9.37 \\ \hline
Human Annotation          & 23.43$^\ast$ & 9.71$^\ast$ & 22.17 & 10.87$^\ast$ & 23.90$^\ast$ & 10.09$^\ast$ & 22.75$^\ast$ & 11.04$^\ast$       \\
\Xhline{4\arrayrulewidth}
\end{tabular}
}
\caption{Against the original subject as reference.
}
\label{tab:result_orig}
\end{subtable}

\begin{subtable}{1\textwidth}
\resizebox{\textwidth}{!}{
\begin{tabular}{l|cccc|cccc}
\Xhline{4\arrayrulewidth}
                          & \multicolumn{4}{c|}{Dev} & \multicolumn{4}{c}{Test}\\
                          & R-1 & R-2 & R-L & METEOR & R-1 & R-2 & R-L & METEOR \\ \Xhline{4\arrayrulewidth}
LEAD-2                    & 18.88 & 9.47 & 17.41 & \underline{20.70} & 18.29 & 8.54 & 16.62 & \textbf{20.23} \\
TextRank                  & 18.29 & 8.04 & 16.45 & 17.00$^\ast$ & 17.93 & 7.47 & 16.00 & 16.98$^\ast$  \\
LexRank                   & 21.82 & \underline{10.83}$^\ast$ & 19.78 & \textbf{20.82} & 20.84 & \underline{9.57}$^\ast$ & 18.68 & \underline{19.97} \\
\newcite{shang2018unsupervised}    & 16.28 & 6.14 & 15.07 & 12.12 & 16.11 & 5.50 & 14.88 & 11.81\\ 
\newcite{see2017get}      & \underline{23.37} & 7.36 & \underline{20.99} & 16.27$^\ast$ & \underline{23.31} & 7.28 & \underline{20.83} & 15.68$^\ast$ \\
\newcite{paulus2018deep}  & 15.12 & 4.62 & 13.98 & 10.82 & 14.56 & 4.39 & 13.53 & 10.37  \\
\newcite{hsu2018unified}  & 22.98 & 7.07 & 19.95 & 18.83 & 22.80 & 7.09 & 19.85 & 18.45  \\
\newcite{narayan2018dont} & 11.33 & 1.45 & 11.14 & 4.90  & 11.53 & 1.37 & 11.40 & 5.04 \\
Our System                & \textbf{25.39} & \textbf{10.94} & \textbf{24.72} & 13.04 & \textbf{26.11} & \textbf{11.43} & \textbf{25.64} & 13.52 \\ \hline
Original Subject          & 24.38$^\ast$ & 10.15$^\ast$ & 23.00$^\ast$ & 16.49$^\ast$ & 24.57 & 10.40 & 23.15 & 14.08       \\
Human Annotation          & 35.93 & 17.76 & 33.55 & 21.74 & 36.19 & 17.75 & 33.50 & 21.42       \\ \Xhline{4\arrayrulewidth}
\end{tabular}
}
\caption{Against two human annotations as reference.}
\label{tab:result_human}
\end{subtable}

\caption{Automatic metric scores. \textbf{bold}: best. \underline{underlined}: second best.
$^\ast$ indicates there is no statistically significant difference from our system with $p < 0.01$ under a paired t-test.
}
\vspace{-2mm}
\label{tab:result_auto}
\end{table*}

%
%
%

\section{Experimental Setup}
\subsection{Evaluation}
\label{sec:metric}
\noindent \textbf{Automatic Evaluation.}
Since SLG is a new task, we analyze the usefulness of automatic metrics from sister tasks, and also use human evaluation.
We first use automatic metrics from text summarization and machine translation: (1) ROUGE \cite{lin2004rouge} including F1 scores of ROUGE-1, ROUGE-2, and ROUGE-L. (2) METEOR \cite{denkowski2014meteor}.
They all rely on one or more references and measure the similarity between the output and the reference.
In addition, we include \esqe, which is a reference-less metric.

\noindent \textbf{Human Evaluation.}
While those automatic scores are quick and inexpensive to calculate, only our quality estimator is designed for evaluation of subject line generation.
Therefore, we also conduct an extensive human evaluation on the \textit{overall} score and two aspects of email quality: informativeness and fluency.
An email subject is \textit{informative} if it contains accurate and consistent details with the body, and it is \textit{fluent} if free of grammar errors.
We show the email body along with different system outputs as potential subjects (the models are anonymous).
For each subject and each aspect, the human judge chooses a rating from 1 for Poor, 2 for Fair, 3 for Good, 4 for Great.
We randomly select 500 samples and have each rated by 3 human judges.

\subsection{Baselines}
To benchmark our method, we use several methods from the summarization field, including some recent state-of-the-art systems, because the email subject line can be viewed as a short summary of the email content.
They can be clustered into two groups.\\
\noindent \textbf{(1) Unsupervised extractive or/and abstractive summarization.}
\textbf{LEAD-2} directly uses the first two sentences as the subject line. We choose lead-2 to include both the greeting and the first sentence of main content.
\textbf{TextRank} \cite{mihalcea2004textrank} and \textbf{LexRank} \cite{erkan2004lexrank} are two graph-based ranking models to extract the most salient sentence as the subject line.
\textbf{\newcite{shang2018unsupervised}} use a graph-based framework to extract topics and then generate a single abstractive sentence for each topic under a budget constraint.
\\
\noindent \textbf{(2) Neural summarization using encoder-decoder networks with attention mechanisms.} \cite{sutskever2014sequence,bahdanau2015neural}.
The Pointer-Generator Network from \textbf{\newcite{see2017get}} augments the standard encoder-decoder network by adding the ability to copy words from the source text and using the coverage loss to avoid repetitive generation.
\textbf{\newcite{paulus2018deep}} propose neural intra-attention models with a mixed objective of supervised training and policy learning.
\textbf{\newcite{hsu2018unified}} extend the pointer-generator network by unifying the sentence-level attention and the word-level attention.
\textbf{\newcite{narayan2018dont}} use a topic-based convolutional neural network to generate extreme summarization for news documents.
While they are quite successful in single document summarization, they are mostly extractive, exhibiting a small degree of abstraction \cite{narayan2018dont}.
It is unclear how they perform to generate email subject lines of extremely abstractive summarization.
We train these models on our dataset.

\subsection{Implementation Details}
\textbf{Our Model}.
We pretrain 128-dimensional word2vec \cite{mikolov2013distributed} on our corpus as initialization and update word embeddings during training.
We use single layer bidirectional LSTMs with 256 hidden units in all models.
The convolutional sentence encoders have filters with window sizes (3,4,5) and there are 100 filters for each size.
The batch size is 16 for all training phases.
We use the Adam optimizer \cite{kingma2015adam} with learning rates of 0.001 for supervised pretraining and 0.0001 for RL.
We apply gradient clipping \cite{pascanu2013difficulty} with L2-norm of 2.0.
The training is stopped early if the validation performance is not improved for 3 consecutive epochs.
All experiments are performed on a Tesla K80 GPU.
All submodels can converge within 1-2 hours and 10 epochs so the whole training takes about 4 hours.\\

\textbf{Baselines}.
For \textbf{TextRank} and \textbf{LexRank}, we use the sumy\footnote{\url{https://github.com/miso-belica/sumy}} implementation which uses the snowball stemmer, the sentence and word tokenizer from NLTK\footnote{\url{https://www.nltk.org/}}.
For \textbf{\newcite{shang2018unsupervised}}, we use their extension of the Multi-Sentence Compression Graph (MSCG) of \newcite{filippova2010multi} and a budget of 10 words in the submodular maximization.
We choose the number of communities from [1,2,3,4,5] based on the dev set and we find that 1 works best. 
For the Pointer-Generator Network from \textbf{\newcite{see2017get}}, we follow their implementation\footnote{\url{https://github.com/abisee/pointer-generator}} and use a batch size 16.
For \textbf{\newcite{paulus2018deep}}, we use an implementation from \newcite{keneshloo2018deep}\footnote{\url{https://github.com/yaserkl/RLSeq2Seq}}.
We did not include the intra-temporal attention and the intra-decoder attention because they hurt the performance.
For \textbf{\newcite{hsu2018unified}}, we follow their code\footnote{\url{https://github.com/HsuWanTing/unified-summarization}} with a batch size 16.
All training is early stopped based on the dev set performance.

%
%

\section{Results and Discussion}
\subsection{Automatic Metric Evaluation}
We report the automatic metric scores against the original subject and the subjects generated by Turkers (human annotations) as references in Tables \ref{tab:result_orig} and  \ref{tab:result_human} respectively.
Table \ref{tab:result_estimator} also shows the \esqe scores.
Overall, our method outperforms the other baselines in all metrics except METEOR.
Other systems can achieve higher METEOR scores because METEOR emphasizes recall (recall weighted 9 times more than precision) and other extractive systems such as LexRank can generate longer sentences as subject lines.

In Table \ref{tab:result_orig}, where the original subject is the singular reference, the score of our system is rated close to and even higher than the human annotation on both sets.
This is because our system is trained on the original subject and is likely a better domain fit.
In Table \ref{tab:result_human}, all systems use two human annotations as the reference to have a fair comparison to the human-to-human agreement in the last row.
Our system output is actually rated a bit higher than the original subject.
This is because the original subject can differ from the human annotation when the sender and the recipient share some background knowledge hidden from the email content.
Furthermore, in the last row, the human-to-human agreement is much higher than all the system outputs and the original subject.
This indicates that different annotators write subjects with a similar choice of words.
In Table \ref{tab:result_estimator}, \esqe still considers our system better than other baselines, while the human annotation has the best quality score.\\
\textbf{Evaluation of sub-components}.
Our extractor captures salient information by selecting multiple sentences from the email body.
We measure its performance as a classification problem against the ``proxy" sentence labels as explained in Section \ref{sec:training}.
The overall precision and recall on the test set is 74\% and 42\%, respectively.
Out of 1906 test examples, 691 examples have more than one sentence selected, and 1626 first sentences and 973 non-first sentences are extracted.
Furthermore, during RL training phase, the dev \esqe score increases from 2.30 to 2.40.

\begin{table}[t!]
\centering
\resizebox{\columnwidth}{!}{
\begin{tabular}{l|cc}
\Xhline{4\arrayrulewidth}
                          & Dev  & Test \\ \Xhline{4\arrayrulewidth}
LEAD-2                    & 1.56 & 1.55 \\
TextRank \cite{mihalcea2004textrank}                 & 1.59 & 1.59 \\
LexRank  \cite{erkan2004lexrank}                 & 1.57 & 1.56 \\
\newcite{shang2018unsupervised}    & 2.10 & 2.09 \\ 
\newcite{see2017get}      & 2.22 & 2.19 \\
\newcite{paulus2018deep}  & \underline{2.30} & \underline{2.30} \\
\newcite{hsu2018unified}      & 1.44 & 1.46 \\
\newcite{narayan2018dont}  & 1.53 & 1.54 \\
Our System                & \textbf{2.40} & \textbf{2.39} \\
\hline
Original Subject          & 2.52 & 2.51 \\
Human Annotation          & 2.53 & 2.54 \\ \Xhline{4\arrayrulewidth}
\end{tabular}
}
\caption{\esqe score. Compared with our system, all other are statistically significant with $p<0.01$ under a paired t-test.}
\label{tab:result_estimator}
\end{table}

\begin{table}[t!]
\centering
\resizebox{\columnwidth}{!}{
\begin{tabular}{lccc}
\Xhline{4\arrayrulewidth}
& Overall & Informative & Fluent \\
\Xhline{4\arrayrulewidth}
Random                 & 1.10$^{\ast}$  & 1.45 & 2.21 \\
\newcite{see2017get}   & 1.45$^{\ast}$  & 1.98 & 1.61 \\
Our System             & \textbf{2.28}  & \textbf{2.38} & \textbf{2.89} \\ \hline
Original Subject       & 2.56  & 2.66 & 3.11 \\
Human Annotation       & 2.74$^{\ast}$  & 3.07 & 2.94\\
\Xhline{4\arrayrulewidth}                  
\end{tabular}
}
\caption{Human evaluation.
$^{\ast}$ indicates the difference from our system is statistically significant with $p < 0.01$ under a paired t-test.
}
\label{tab:human_eval}
\end{table}

\begin{table}[t!]
\centering
\resizebox{\columnwidth}{!}{
\begin{tabular}{lcc}
\Xhline{4\arrayrulewidth}
                              & Pearson's $r$ & Spearman's $\rho$      \\ \Xhline{4\arrayrulewidth} 
\esqe                         & \textbf{0.49} & \textbf{0.46} \\ 
ROUGE-1 F1                    & 0.44 & 0.43 \\
METEOR                        & 0.40 & 0.40 \\
\hline
Inter-Rater Agreement         & 0.64 & 0.58 \\ \Xhline{4\arrayrulewidth} 
\end{tabular}
}
\caption{Correlation analysis between the automatic scores and the human evaluation.}
\label{tab:correlation}
\end{table}

\begin{table*}[t!]
\centering
\begin{subtable}{1\textwidth}
\centering
\resizebox{\textwidth}{!}{
\begin{tabular}{?p{18cm}?}
\Xhline{4\arrayrulewidth}
\textbf{Email Body:} Dear Rick, Thanks for speaking with me today.
\ul{Here is the position description for the KWI President of the Americas Opportunity.}
\ul{I feel that this is a tremendous opportunity to be an integral player with a very exciting relatively early stage Applications Software  company, in the very exciting and hot Energy Commodities Sector;  They are already profitable, pre-IPO.}
This position has a great compensation package.
Please get back to me if you have an interest or if you know someone who might be intrigued by this opportunity.
Thanks, Dal Coger \\
\textbf{Original Subject:} KWI President of the Americas \\
\textbf{Human Annotation:} KWI President of the Americas Opportunity\\
\textbf{See et al., ACL 2017:} Position Description - the Americas Sector Opportunity \\
\textbf{Our System:} KWI President of the Americas Position \\ \Xhline{4\arrayrulewidth}
\end{tabular}
}
\caption{\textbf{Email ID:} buy-r\_inbox\_321}
\label{tab:ex1}
\end{subtable}

\begin{subtable}{1\textwidth}
\centering
\resizebox{\textwidth}{!}{
\begin{tabular}{?p{18cm}?}
\Xhline{4\arrayrulewidth}
\textbf{Email Body:}  \ul{Attached for your information are the following two filings made at FERC on Monday on behalf of WPTF:   1..  Motion to Intervene and Protest of the Western Power Trading Forum.}
\ul{This was filed in connection witht the ISO status report filing dealing with creditworthiness issues.}
2..  Motion to Intervene and Comments of the Western Power Trading Forum.
This was filed in connection with the Reliant and Mirant filing of a joint Section 206 complaint on October 18, 2001.
My thanks to those who responded to the drafts with comments and suggestions.
Dan\\
\textbf{Original Subject:} Monday's FERC Filings \\
\textbf{Human Annotation:} Two Filings Made at FERC \\
\textbf{See et al., ACL 2017:}  FERC filings - FERC power and at monday was filing \\
\textbf{Our System:}  Western Power Trading Filings \\
\Xhline{4\arrayrulewidth}
\end{tabular}
}
\caption{\textbf{Email ID:} dasovich-j\_inbox\_1473}
\label{tab:ex2}
\end{subtable}

\begin{subtable}{1\textwidth}
\centering
\resizebox{\textwidth}{!}{
\begin{tabular}{?p{18cm}?}
\Xhline{4\arrayrulewidth}
\textbf{Email Body:} \ul{Hi Evening MBA students,  If you plan to graduate this semester for a December 2001 degree, will you  please come by the Evening MBA office soon (by Tuesday, September 25 at the  latest) and fill out an Application for Candidacy form?}
\ul{We have your fall transcript to assist you in filling out the form.}
Since we need your original signature, an office visit is best.
Thanks, congratulations, and see you!\\
\textbf{Original Subject:} Planning to graduate this semester? \\
\textbf{Human Annotation:}  December 2001 degree \\
\textbf{See et al., ACL 2017:} December application(graduate) - September 25 \\
\textbf{Our System:} December 2001 degree application \\
\Xhline{4\arrayrulewidth}
\end{tabular}
}
\caption{\textbf{Email ID:} dasovich-j\_inbox\_123}
\label{tab:ex3}
\end{subtable}

\begin{subtable}{1\textwidth}
\centering
\resizebox{\textwidth}{!}{
\begin{tabular}{?p{18cm}?}
\Xhline{4\arrayrulewidth}
\textbf{Email Body:} \ul{As our last day is Friday, November 30th, we would love to toast the good times and special memories that we have shared with you over the past five years.}
\ul{Please join us at Teala's (W. Dallas) on Thursday, November 29th, beginning at 5pm.}
Looking forward to being with you,   Lara and Janel     Lara Leibman
\\ 
\textbf{Original Subject:} Farewell Drinks \\
\textbf{Human Annotation:} Our last day \\
\textbf{See et al., ACL 2017:} Friday 30th and day, W. Dallas - November \\
\textbf{Our System:} Teala's \\ \Xhline{4\arrayrulewidth}
\end{tabular}
}
\caption{\textbf{Email ID:} arnold-j\_inbox\_153}
\label{tab:ex4}
\end{subtable}
\caption{Case study. The sentences extracted by our model are underlined. 
(a)(b)(c): Our model can generate effective subjects by extracting and rewriting multiple sentences containing salient information. 
(d): Our model fails to generate reasonable subjects for the novel topic of ``farewell" which is not seen during training.
}
\label{tab:case_study}
\end{table*}

\subsection{Human Evaluation}
Table \ref{tab:human_eval} shows that our system is rated higher than the baselines on overall, informative, and fluent aspects.
For overall scores, the baselines are all between 1.5 and 2.0, indicating the subjects are usually considered as poor or fair  (recall that the scale is 1-4, with 4 being the highest).
Our system is 2.28, while the original subject and human annotation are between 2.5 and 3.0.
This means more than half of our system outputs are at least fair, and the original subject and human annotation are often good or great.
We also find that in 89 out of 500 emails, our system outputs have ratings higher than or equal to the original and human annotated subjects.  
Furthermore, the raters prefer the human annotated subject to the original subject.

\subsection{Metric Correlation Analysis}
It is important to check if the automatic metric scores can truly reflect the generation quality and serve as valid metrics for subject line generation.
Therefore, in Table \ref{tab:correlation}, we investigate their correlations with the human evaluation.
To this end, we take the average of three human ratings and then calculate Pearson's $r$ and Spearman's $\rho$ between different automatic scores and the average human rating.
We also report the inter-rater agreement in the last row by checking the correlation between the third human rating and the average of the other two.
We find that the inter-rater agreement is moderate with 0.64 for Pearson's $r$ and 0.58 for Spearman's $\rho$.
We would recommend \esqe because it has the highest correlations while being reference-less.

\subsection{Case Study}
\label{sec:case_study}
Table \ref{tab:case_study} shows examples of our model outputs.
Our model works well by first picking multiple sentences containing information such as named entities and dates and then rewriting them into a succinct subject line preserving the key information.
In Example \ref{tab:ex1}, our model extracts sentences with the name of the company and position ``KWI President of the Americas".
It also captures the importance of the opportunity for this position.
Similarly, in Example \ref{tab:ex2}, our model identifies ``Western Power Trading" for ``filings".
In Example \ref{tab:ex3}, our model identifies the date of degree ``December 2011" and action item ``application".
However, we also found our model can fail on emails about novel topics, as in Example \ref{tab:ex4} where the topic is scheduling farewell drinks.
Our model only captures the name of the restaurant but not the purpose and topic since it has not seen this kind of email in training.

\section{Related Work}
\vspace{-1mm}
Past NLP email research has focused on summarization \cite{muresan2001combining,nenkova2003facilitating,rambow2004summarizing,corston2004task,wan2004generating,carenini2007summarizing,zajic2008single,carenini2008summarizing,ulrich2009regression}, keyword extraction and action detection \cite{turney2000learning,bennett2005detecting,dredze2008generating,scerri2010classifying,loza2014building,lahiri2017keyword,lin2018actionable}, and classification \cite{prabhakaran2014gender,prabhakaran2014predicting,alkhereyf2017work}.
However, we could not find any previous work on email subject line generation.
The very first study on email summarization is \newcite{muresan2001combining} who reduce the problem to extracting salient phrases.
Later, \newcite{nenkova2003facilitating}, \newcite{rambow2004summarizing}, \newcite{wan2004generating} deal with the problem of email thread summarization by the sentence extraction approach.

Another related line of research is natural language generation.
Our task is most similar to single document summarization because the email subject line can be viewed as a short summary of the email content.
Therefore, we use different summarization models as baselines with techniques such as graph-based extraction and compression, sequence-to-sequence neural abstractive summarization with the hierarchical attention, copy, and coverage mechanisms.
In addition, RL has become increasingly popular for text generation to optimize the non-differentiable metrics and to reduce the exposure bias in the traditional ``teaching forcing" supervised training \cite{ranzato2016sequence,bahdanau2017actor,zhang2017sentence,sakaguchi2017grammatical}.
For example, \newcite{narayan2018ranking} use RL for ranking sentences in pure extractive summarization.

Furthermore, current methods on news headline generation \cite{lopyrev2015generating,tilk2017low,kiyono2017source,tan2017neural,shen2017recent} most follow the encoder-decoder model, while our model uses a multi-sentence selection and rewriting framework.

\section{Conclusions and Future Work}
\vspace{-1mm}
In this paper, we introduce the task of email subject line generation.
We build a benchmark dataset (AESLC) with crowdsourced human annotations on the Enron corpus and evaluate automatic metrics for this task.
We propose our model of subject generation by Multi-Sentence Selection and Rewriting with Email Subject Quality Estimation Reward.
Our model outperforms several competitive baselines and approaches human-level performance.

In the future, we would like to generalize it to multiple domains and datasets.
We are also interested in generating more effective and appropriate subjects by incorporating prior email conversations, social context, the goal and style of emails, personality, among others.

\section*{Acknowledgements}
We would like to thank Jimmy Nguyen and Vipul Raheja for their help in the data creation process. 
We also thank Dragomir Radev, Courtney Napoles, Dimitrios Alikaniotis, Claudia Leacock, Junchao Zheng, Maria Nadejde, Adam Faulkner, and three anonymous reviewers for their helpful discussion and feedback.

\bibliography{acl2019}
\bibliographystyle{acl_natbib}

\appendix
\section{Amazon Mechanical Turk}
\begin{figure*}[t]
  \centering
  \includegraphics[width=\textwidth]{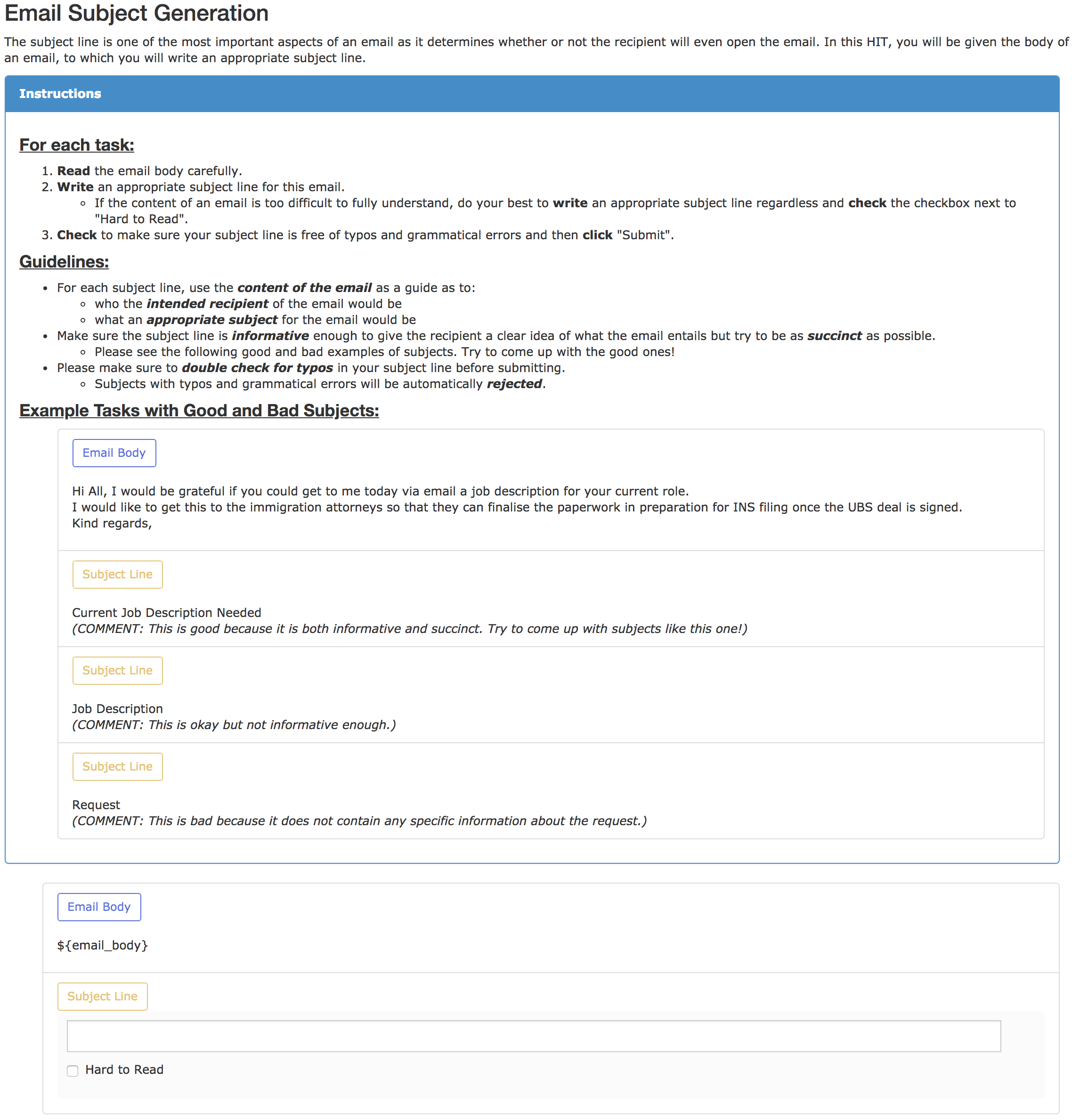}
  \caption{Amazon Mechanical Turk job interface for the email subject annotation.}
  \label{fig:mturk_annotation}
\end{figure*}

\begin{figure*}[t]
  \centering
  \includegraphics[width=\textwidth]{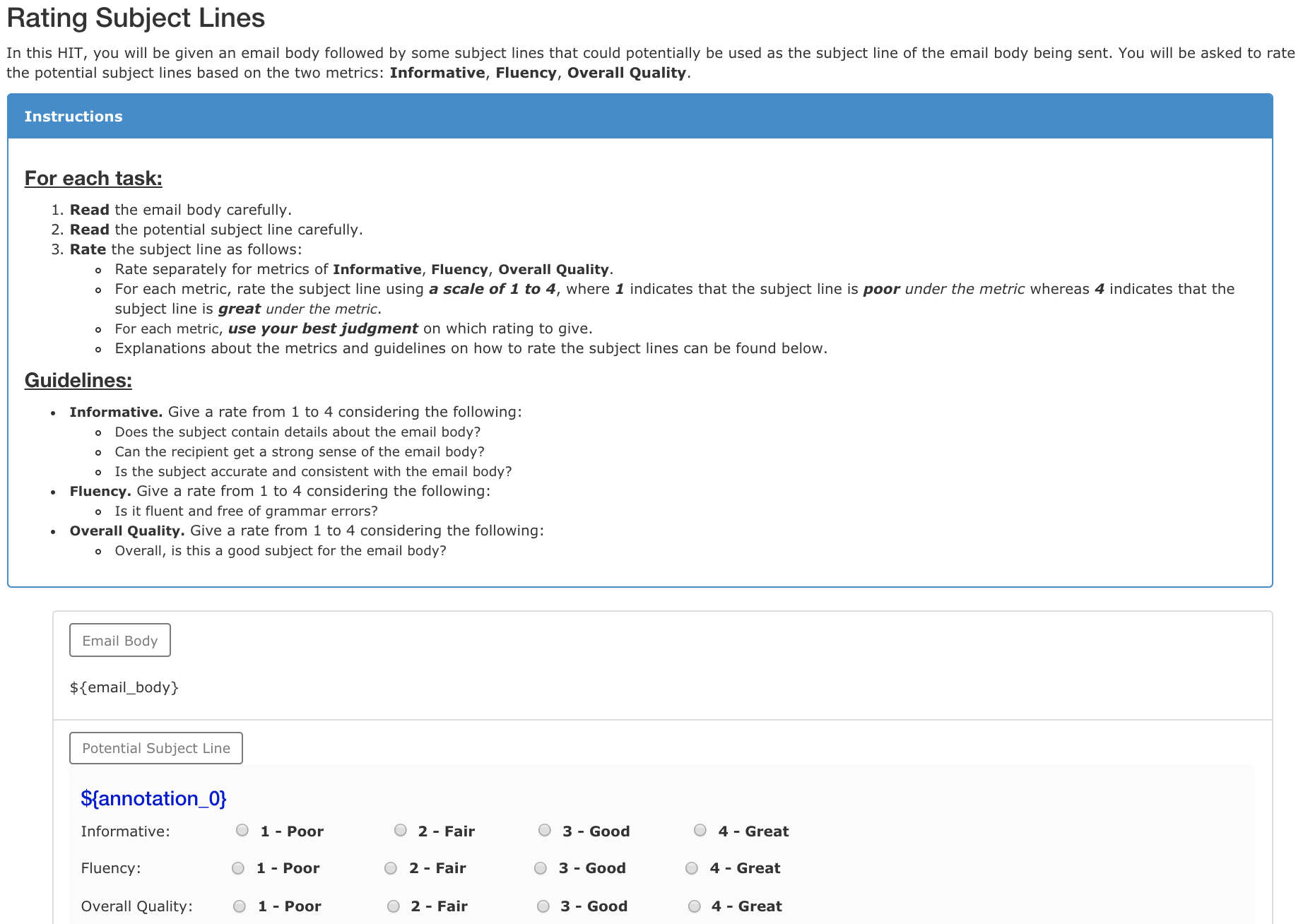}
  \caption{Amazon Mechanical Turk job interface for the email subject evaluation. All the other system outputs are in the same job but are not shown here for brevity.}
  \label{fig:mturk_evaluation}
\end{figure*}

Figure \ref{fig:mturk_annotation} shows the Amazon Mechanical Turk interface for workers to write the email subject from the body.
Figure \ref{fig:mturk_evaluation} shows the interface for the email subject evaluation.
For quality control, we include a random subject.
Annotators who consistently give high ratings for Random subjects or low ratings for Human Annotation subjects are excluded from our analysis.  This filtering resulted in a total of 389 examples with 3 valid ratings each of which we take the average.

\end{document}